\RequirePackage{fix-cm}
\documentclass[twocolumn]{svjour3}          
\smartqed  
\usepackage{amssymb}
\usepackage{booktabs}  
\usepackage{threeparttable}  
\usepackage{multirow}
\usepackage{makecell}
\usepackage{amsmath}
\usepackage{bm}
\usepackage{graphicx}
\usepackage{subfigure}
\usepackage{longtable}
\usepackage{setspace}
\usepackage{epstopdf}
\usepackage{multirow}
\usepackage{changepage}
\usepackage[misc]{ifsym}
\usepackage{amsfonts}
\usepackage{color}
\usepackage{algorithm,algpseudocode}
\usepackage{algorithmicx}
\usepackage{comment}
\usepackage{url}
\begin{document}

\title{Spacecraft Anomaly Detection with Attention Temporal Convolution Networks
}


\author{
        Liang Liu 
        \textsuperscript{1} 
        \and Ling Tian \textsuperscript{1} 
         \and Zhao Kang \textsuperscript{1} 
        \and Tianqi Wan \textsuperscript{2}
}

\institute{
\Letter Zhao Kang\\
\email{Zkang@uestc.edu.cn}\\
 \at
 {1} School of Computer Science and Engineering, University of Electronic Science and Technology of China, Chengdu, China.
 \at
 {2} Beijing Aerospace Institute for Metrology and Measurement Technology, Beijing, China.
}



\date{Received: date / Accepted: date}

\maketitle

\begin{abstract}
Spacecraft faces various situations when carrying out exploration missions in complex space, thus monitoring the anomaly status of spacecraft is crucial to the development of the aerospace industry. The time series telemetry data generated by on-orbit spacecraft contains important information about the status of spacecraft. However, traditional domain knowledge-based spacecraft anomaly detection methods are not effective due to high dimensionality and complex correlation among variables. In this work, we propose an anomaly detection framework for spacecraft multivariate time-series data based on temporal convolution networks (TCNs). First, we employ dynamic graph attention to model the complex correlation among variables and time series. Second, temporal convolution networks with parallel processing ability are used to extract multidimensional features for the downstream prediction task. Finally, many potential anomalies are detected by the best threshold. Experiments on real NASA SMAP/MSL spacecraft datasets show the superiority of our proposed model with respect to state-of-the-art methods. 


\end{abstract}

\section{Introduction}\label{secition1}

Deep space of the solar system has numerous satellites in orbit collecting planetary data for exploration  missions such as Tianwen-1's high-resolution multispectral imagery and magnetic monitoring missions around Mars. These satellites have made great contributions to scientific research and resource and environmental exploration. The internal systems of these spacecraft are composed of various sophisticated technologies such as telemetry sensing, navigation control, and many others, which make the system structure complicated \cite{zhang2019contribution}. Furthermore, spacecraft operate in an extremely complex deep space environments and are confronted with unforeseen anomalies and failures. Therefore, it is vital to effectively and timely detect anomalies in components of spacecraft to ensure its reliable, safe, and continuous operation during exploration missions \cite{chen2021imbalanced}.  

For on-orbit spacecraft, the most commonly used method is to collect real-time operation data of each component from multi-sensors \cite{jiang2022anomaly} to monitor the status of internal spacecraft. These data are converted into electrical signals and transmitted to the ground telemetry center, where the original variable information of each channel is restored through the signal demodulation technology \cite{wang2021anomaly}. Then, pattern discovery and anomaly detection analysis can be performed on such telemetry data, which are multidimensional time series \cite{chalapathy2019deep,hundman2018detecting}. In fact, spacecraft telemetry anomaly detection is an intractable problem. There are thousands of telemetry channels that need to be monitored and data have sophisticated patterns. It is impossible for domain experts to observe each channel and manually mark anomalies in a predefined range \cite{chang1992satellite}.  

Many data-driven anomaly detection methods have been proposed for multivariate time series data \cite{wang2019multivariate}. These approaches build a mathematical model for spacecraft normal pattern from telemetry data to detect anomalies without using any prior knowledge of experts. 
Some representative statistical models are autoregressive moving average (ARMA) \cite{galeano2006outlier}, gaussian mixture \cite{li2016anomaly},  and autoregressive integrated moving average (ARIMA) \cite{zhang2005network}. However, these models perform poor facing huge, non-linear and high-dimensional time series data. Inspired by the success of deep learning, 
many anomaly detection methods are built upon deep neural networks \cite{choi2021deep,mathonsi2022multivariate}. Due to the lack of labels, most methods follow an unsupervised learning scheme \cite{shi2022unsupervised}, which learns normal and expected behavior of telemetry channel by predicting \cite{ding2019real,hundman2018detecting} or reconstructing expected errors \cite{zhang2019deep,kang2019robust}. They use some popular architectures, including long short-term memory networks (LSTMs) \cite{hsieh2019unsupervised,park2018multimodal}, auto-encoders (AE) \cite{audibert2020usad,wen2019time,su2019robust}, generative adversarial networks (GANs) \cite{li2019mad,zhou2019beatgan,choi2020gan}, and Transformer \cite{chen2021learning,meng2019spacecraft}. 
These deep learning methods have achieved significant performance improvements for time-series anomaly detection. However, there are several downsides in applying them to spacecraft telemetry data. First, most existing methods need to construct a separate model to monitor each telemetry channel, which fails to consider the potential correlations in real spacecraft datasets. Second, they require long training time to support complex computation.

To this end, we propose an anomaly detection framework for spacecraft multivariate time series data based on temporal convolution networks
(TCNs). Spacecraft data including the Soil Moisture Active Passive (SMAP) satellite and the Mars Science Laboratory (MSL) rover \cite{hundman2018detecting} are applied to verify the effectiveness of our proposed framework. Specifically, the main contributions are:
\begin{itemize}
    \item First, we exploit dynamic graph attention to model the complex correlation among variables and capture the long-term relationship. 
    \item Second, the TCNs with parallel processing ability is used to extract multidimensional features. 
    \item Third, the static threshold method is applied to detect potential anomalies in real-world spacecraft data. 
\end{itemize}

\section{Related Work}\label{secition2}

\begin{figure}[htp]
	\centering
	\includegraphics[width=0.48\textwidth]{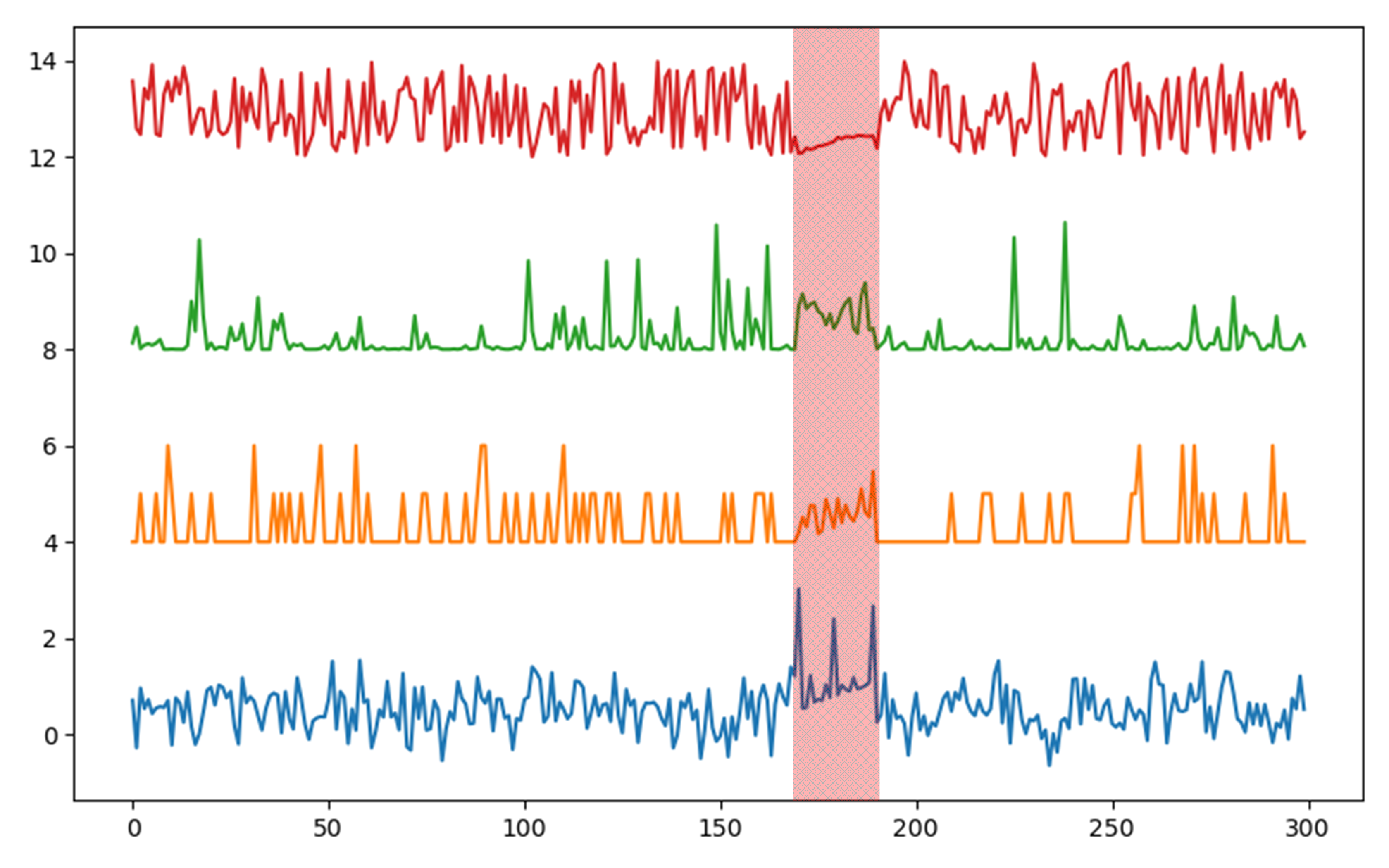}
	\caption{Illustrative example of anomalies in telemetry data.}
	\label{anomaly-example}
\end{figure}

Recently, deep neural network architectures have achieved leading performance on various time series anomaly detection tasks. 
Especially, LSTMs and RNNs can effectively process time-series data, capturing valuable historical information for future prediction. Based on LSTMs, NASA Jet Propulsion Laboratory designs an unsupervised nonparametric anomaly thresholding approach for spacecraft anomaly detection \cite{hundman2018detecting}; Park \emph{et al}. \cite{park2018multimodal} propose variational auto-encoders that fuse signals and reconstruct expected distribution. Most telemetry data are multi-dimensional due to the interrelation of components in the satellite's internal structure. As illustrated in Fig. \ref{anomaly-example}, each curve corresponds to a variable (or channel) in 
spacecraft multivariate time series, and an anomaly in one channel of the spacecraft also causes abnormality in other channels.
For online anomaly detection task, they have to create a model for each variable and simultaneously invoke multiple trained models, which are inefficient. Therefore, the above methods cannot detect anomalies due to the complex correlations among multivariate. 

Consequently, the methods considering multivariable and their potential correlations have received extensive attention. For example, NASA Ames Research Center proposes a clustering-based inductive monitoring system (IMS) to analyze archived system data and characterize normal interactions between parameters \cite{iverson2012general}. 
Su \emph{et al}. \cite{su2019robust} utilize stochastic variable connection and planar normalizing flow for multivariate time series anomaly detection, which performs well in datasets of NASA. 
Zhao \emph{et al}. \cite{zhao2020multivariate} utilizes the graph attention method to capture temporal and variable dependency for anomaly detection. Li \emph{et al}. \cite{li2018anomaly,li2019mad} adopts GANs to detect anomalies by considering complex dependencies between multivariate. Zong \emph{et al}. \cite{zong2018deep} introduce deep auto-encoders with a gaussian mixture model.
Chen \emph{et al}. \cite{chen2021learning} directly built a transformer-based architecture that learns the inter-dependencies between sensors for anomaly detection.

Although these methods have made a significant improvement in anomaly detection, there are still some shortcomings. For example, RNNs processing the next sequence need to wait for the last output of the previous sequence. Therefore, modeling long-time series data requires a long time, which leads to inefficient detection and does not meet the real-time requirement for spacecraft telemetry anomaly detection. Additionally, most models have high complexity and are not suitable for the practical application of spacecraft anomaly detection. Bai \emph{et al}. \cite{bai2018empirical} firstly propose TCNs that have parallel processing ability and also model historical information with exponential growth receptive field. In recent years, many studies have verified the effectiveness of TCNs for long time series \cite{he2019temporal}. However, anomaly detection based on TCNs has not been investigated. Therefore, this paper develops a multivariate temporal convolution anomaly detection model for complex spacecraft telemetry data.

\begin{figure*}[htp]
	\centering
	\includegraphics[width=\textwidth]{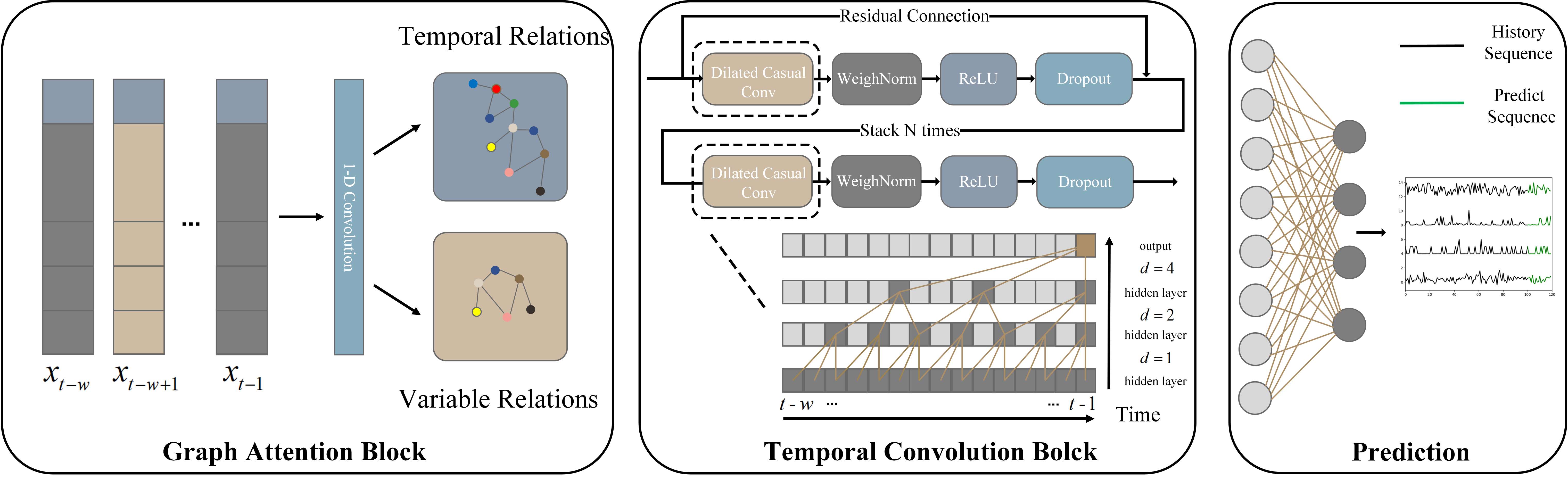}
	\caption{The overview of our proposed architecture for spacecraft anomaly detection. }
	\label{model}
\end{figure*}

\section{Methodology}\label{secition3}
The proposed spacecraft multivariate telemetry data anomaly detection method aims to detect anomalies at entity-level \cite{su2019robust} instead of channel-level since the overall status is more noteworthy and less expensive to observe. 
Let $X\in \mathcal{R}^{w \times m}$ denotes the multivariate time series, where $w$ is the length of input timestamp and $m$ is the number of feature (or called variable) of input. 
Given a historical sequence $\{x_{t-w},...,x_{t-1} \}  $, our goal is to predict its value at the next time step $t$.
Based on it, we can calculate the anomaly score  and threshold, and finally output  $y_t \in \{0, 1\}$, indicating whether it is anomalous or not at timestamp $t$. Considering the imbalance between anomalous and normal spacecraft telemetry data, we set the model to learn a normal mode by offline training and to monitor the anomalies online.

The overview of our proposed spacecraft anomaly detection architecture is shown in Fig. \ref{model}, which consists of three modules:
(a) Given a multivariate time-series sequence, we apply 1-D convolution to extract high-level features and use dynamic attention to capture temporal and variable relations. (b) Concatenate the outputs of the dynamic attention layer and convolution layer, and perform multi-stack temporal convolution to encode multivariate time-series representation. c) Multi layered perceptron (MLP) is used to map the encoding representation to predict future values. We minimize the residuals between predicted and observed values to update the model until convergence.

\subsection{Graph Attention Mechanism}
As aforementioned, spacecraft multivariate telemetry time series have complex interdependence. We use graph structure to model the relationships between variables \cite{fang2022structure}. Given an undirected graph $\mathcal{G}=(\mathcal{V, \mathcal{E}})$, where $\mathcal{V}$ denotes the set of vertices (or nodes) and $\mathcal{E} \subseteq {\mathcal{V} \times \mathcal{V}}$ is a set of edges.
Multivariate time series $X = [x_{t-w},...,x_{t-1}] \in \mathcal{R}^{w \times m}$ and $x_{i}=\{x_i^{(1)},x_i^{(2)},...,x_i^{(m)} \} \in \mathcal{R}^m $, where $i$ denotes the $i$-th time step (or node $i$), which can be considered as $w$ nodes and $x_{i}$ is feature vector corresponding to each node. It can be used to capture temporal dependencies. 
At the same time, to sufficiently capture the relationship of variables, the feature  matrix can be transformed into $X^{{\rm T}}=[x^{(1)},...,x^{(m)}] \in \mathcal{R}^{m \times w}$, $x^{(i)} = \{x^{(i)}_{t-w}, x^{(i)}_{t-w+1},...,x^{(i)}_{t-1}\} \in \mathcal{R}^w$.

\begin{figure}[htp]
	\centering
	\includegraphics[width=0.48\textwidth]{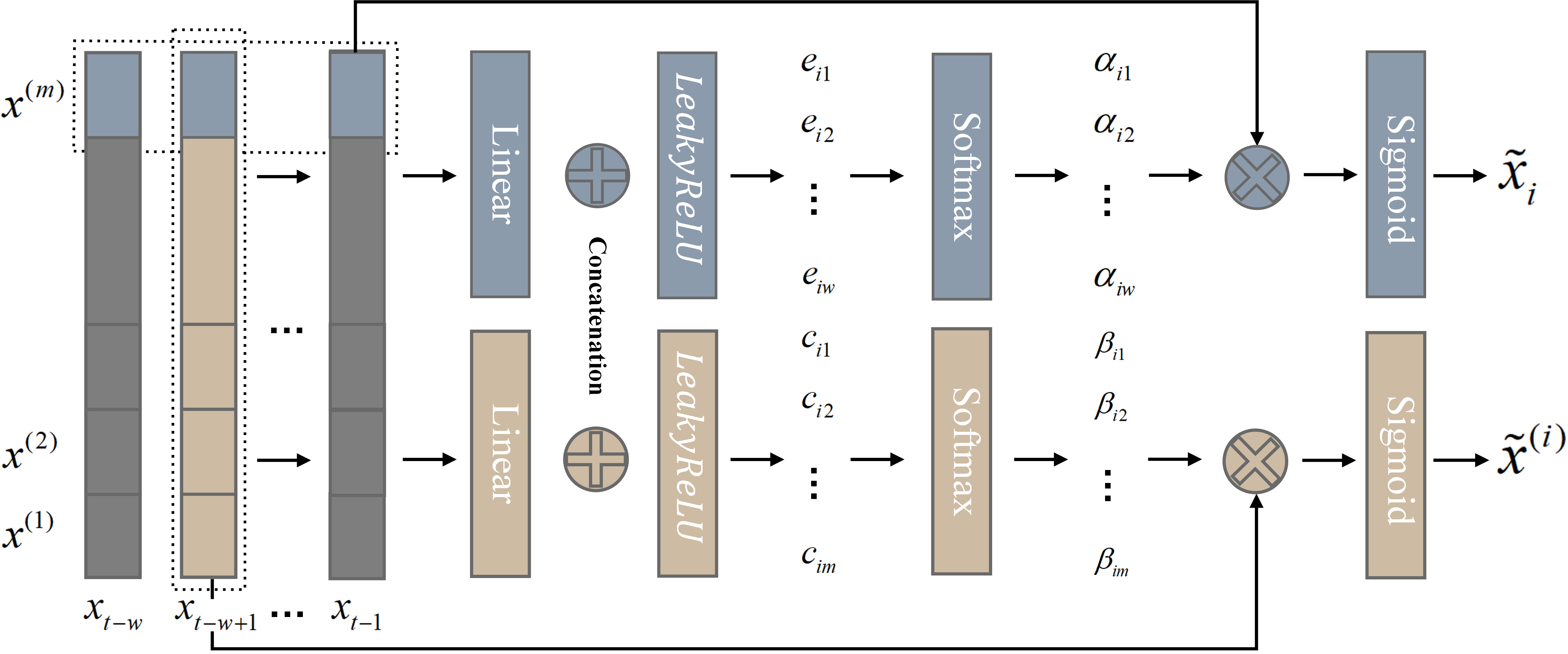}
	\caption{The illustration of temporal attention and variable attention. }
	\label{attention}
\end{figure}

We capture temporal and variable dependencies according to its importance in the fully connected graph as shown in Fig. \ref{attention}. Specifically, graph attention block is based on graph attention networks (GATs) \cite{velivckovic2017graph}. Taking temporal attention for example, 
it learns a weighted averaged of the representation of neighbor nodes as follows:

\begin{equation}
	\widetilde{x}_{i} = \sigma(\sum_{j \in \mathcal{N}_{i}} \alpha_{ij} x_{j} )
\end{equation}
where $\widetilde{x}_{i}$ denotes the output representation of temporal attention on $x_{i}$, $\sigma$ is a nonlinear activation function, and  the attention score $\alpha_{ij}$ is normalized across all  neighbors $j \in \mathcal{N}_i$ by softmax.

\begin{equation}
	\alpha_{ij}=softmax(e(x_i,x_j))=\frac{exp(e(x_i, x_j))}{\sum_{k \in \mathcal{N}_{i}}exp(e(x_i, x_{k}))}
\end{equation}
where scoring function $e(x_i, x_j)$ is defined as:
\begin{equation}
	e_{ij}=e(x_i,x_j)=LeakyReLU(\textbf{a}^{\rm{T}} \cdot [\textbf{W}x_{i}||\textbf{W}x_{j}])    
\end{equation}
which measures the importance of features of neighbor $j$ to node $i$, $||$ denotes concatenation operation,
$\textbf{a}$ and $\textbf{W}$ are trainable parameters. In fact, GATs are static attention and can deteriorate the model fitting capability \cite{brody2021attentive}. This is because the learned layers $\textbf{a}$ and $\textbf{W}$ are applied consecutively, and thus can be collapsed into a single linear layer. We apply dynamic graph attention by changing the order of operations in GATs. Concretely, after concatenating, a linear transformation is applied and then an attention layer is utilized after the activation function. Mathematically, dynamic attention can be defined as below:

\begin{equation}
	e_{ij}=e(x_i, x_j) = \textbf{a}^{\rm{T}}LeakyReLU(\textbf{W} \cdot [x_{i}||x_{j}])
\end{equation}

\subsection{Temporal Convolution Encoding}
As aforementioned, existing methods do not consider the model complexity and long training time. TCNs have proved to be superior to LSTMs and RNNs in terms of computational speed and ability to mine historical information for very long time series.
We utilize TCNs as a backbone network to learn the representation of time series. TCNs are an improved architecture that overcomes the limitations of RNNs-based models, which are not able to capture the property of long time series and have low computational efficiency. TCNs' parallel structure makes it appealing to boost the efficiency of spacecraft anomaly detection. TCNs use a 1-D fully-convolutional network as architecture and zero padding to guarantee the hidden layer be the same size as the input, and also apply causal convolution. 

Suppose convolution filter $F=\{f_1, f_2, ..., f_K\}$, where $K$ denotes the size of convolution kernel. 
The mathematical of the element $x_{t-1}$ by causal convolution can be defined as follow:
\begin{equation}
	F(x_{t-1}) = \sum^{K}_{k=1}f_{k}x_{t-1-K+k}
\end{equation}

It can be observed from the above equation that the ability to process long sequence data is limited unless a large number of layers are stacked, which will make the task inefficient due to limited computing resources. The dilated convolution enlarges the receptive field with limited layers so that each convolution output contains a wide range of information. Dilated convolution reduces the depth of a simple causal convolution network. It also ensures that the output and input size are the same by skipping the input value in a given time step. As the network deepens, its receptive field covering each input in history expands exponentially. The dilated convolution with dilated factor $d=1,2,4$ is shown in Fig. \ref{model} and the mathematical expression can be formulated as:
\begin{equation}
	F_{d}(x_{t-1})=\sum^{K}_{k=1}f_{k}x_{t-1-(K-k)d}
\end{equation}
Finally, the residual connection is applied, which has been proven to be effective for neural network training.
In addition, skipping connection through across-layer manner is also adopted to fully transmit information and avoid vanishing gradient problem in \textcolor{blue}{a} deep network. The output of residual can be formulated as:

\begin{equation}
	F{(x_{t-1})} = F_{d}(x_{t-1}) + x_{t-1}
\end{equation}

\subsection{Spacecraft Anomaly Detection}
\begin{figure}[htp]
	\centering
	\includegraphics[width=0.48\textwidth]{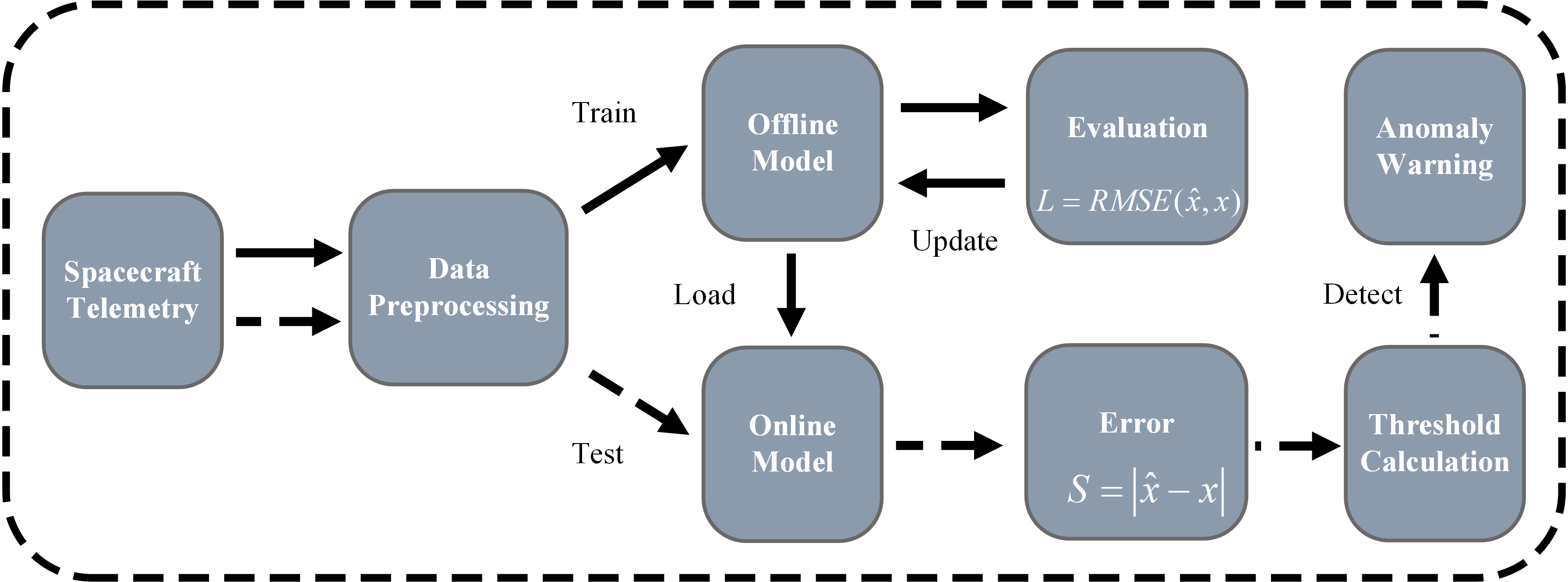}
	\caption{The pipeline of spacecraft anomaly detection.}
	\label{pipline}
\end{figure}

The whole procedure of our proposed Attention Temporal Convolution
Network (ATCN) for spacecraft anomaly detection is shown in Fig. \ref{pipline}. We train the model on normal data to predict telemetry. We minimize the Root Mean Square Error (RMSE) loss $\mathcal{L}=RMSE(\hat{x},x)$ between predicted output sequence $\hat{x}$ and observed sequence $x$, to update the online model. When a detection job arrives, the online model computes the anomalous score sequence $s$ by the deviation level 
between the input sequence and prediction sequence to find the threshold. Then, the error sequence $s$ is compared with the calculated threshold value. If an error at a certain time step exceeds the threshold, the data at this position is considered as an abnormal value.
\section{Experiments}

\subsection{Datasets}
We use real-world spacecraft datasets to verify the effectiveness of our model, including MSL and SMAP \cite{hundman2018detecting}, which are two public datasets of NASA
spacecraft. The statistics of them are shown in Table \ref{tab:datasets}, including the number of variables, size of the training set, size of testing set, the proportion of anomaly in the testing set, and partial variable information.

We first normalize the training and testing dataset. Taking training set as an example, 
\begin{equation}
	x:=\frac{x - \textit{min} (X_{train})}{\textit{max} (X_{train}) - \textit{min} (X_{train})}
\end{equation}
where $\textit{max} (X_{train})$ and $\textit{min} (X_{train})$ are the maximum value and the minimum value of the training set respectively.

\begin{table}[htbp]
	\setlength{\abovecaptionskip}{0pt}%
	\setlength{\belowcaptionskip}{10pt}
	\setlength\tabcolsep{1pt}
	\caption{Dataset statistics}
	\label{tab:datasets}
	\centering
		\begin{tabular}{lcc}
			\hline
			\textbf{\textbf{}} &
			\textbf{SMAP} &
			\textbf{MSL} 
			\\ \hline
			Number of sequences &
			25 &
			55 
			\\ 
			Training set size &
			135183 &
			39312 
			\\
			Testing set size &
			427617 &
			73729 
			\\ 
			Anomaly rate(\%) &
			13.13 &
			10.27 
			\\
			Variable information &
			\multicolumn{2}{l}{\makecell[c]{
					computational,
					radiation, \\temperature, power,
					activities, etc}}
			\\ \hline
		\end{tabular}
\end{table}

\subsection{Baseline}
We compared our model performance with state-of-the-art unsupervised anomaly detection methods, including reconstruction model (R-model) and prediction model (P-model):
\begin{itemize}
	\item KitNet \cite{mirsky2018kitsune} is the first work to use auto-encoder with or without ensembles for online anomaly detection. 
	\item OmniAnomaly \cite{su2019robust} proposes a variational auto-encoder with gated recurrent units for multivariate time series anomaly detection. 
	\item GAN-Li \cite{li2018anomaly} and MAD-GAN \cite{li2019mad} are unsupervised multivariate anomaly detection method based on GANs. They adopt LSTM-RNN as the  generator and discriminator model  to capture the temporal correlation of time series distributions.
    \item LSTM-VAE \cite{park2018multimodal} aims at fusing signals and reconstructing their expected distribution. 
    \item LSTM-NDT \cite{hundman2018detecting} is a univariate time series detection method based on LSTM with a novel nonparametric anomaly thresholding approach for NASA datasets.
    \item DAGMM \cite{zong2018deep} utilizes a deep auto-encoder to generate error sequence and feeds into  Gaussian Mixture Model (GMM) for anomaly detection. 
    \item MTAD-GAT \cite{zhao2020multivariate} captures time relationships and variable dependency with jointly optimizing a forecasting-based model and a reconstruction-based model. 
    \item GTA \cite{chen2021learning} develops a new framework for multivariate time series anomaly detection that involves automatically learning a graph structure, graph convolution, and modeling temporal dependency
	using a Transformer-based architecture.
\end{itemize}
These methods mainly model temporal dependency or multivariate correlation to detect anomalies by reconstruction or prediction model. We implement our method and all its variants with Pytorch version 1.6.0 with CUDA 10.2. The overall experiments are conducted on Tesla T4 GPU, 32G.

\begin{table}
	\caption{Configurations of the model}
	\label{configuration}
	\begin{tabular}{cc}
		\toprule
		Parameters & Configuration\\
		\midrule
		1-D convolution kernel size & 7 \\
		TCNs filter size & 4 \\
		Hidden layers & 2 \\
		Units in hidden layer & 32 \\
		Batch size & 256\\
		Learning rate & 0.001\\
		Dropout & 0.1\\
		Input length (window size) $w$ & 100\\
		Optimizer  & Adam\\
		Forecast loss $\mathcal{L}$ & RMSE\\
		\bottomrule
	\end{tabular}
\end{table}


\begin{table*}[htbp]  
	
	\centering  
	\begin{threeparttable}  
		\caption{Prediction performance of model}  
		\label{tab:performance_comparison}  
		\begin{tabular}{cccccccc}
			\toprule  
			\multicolumn{2}{c}{
				\multirow{3}{*}{Method}}&  
			\multicolumn{3}{c}{SMAP}&\multicolumn{3}{c}{MSL}\cr  
			\cmidrule(lr){3-5} \cmidrule(lr){6-8}  
			&&Precision&Recall&F1&Precision&Recall&F1\cr  
			\midrule  
			\multirow{5}{*}{R-model}& 
			
			\multicolumn{1}{c}{KitNet \cite{mirsky2018kitsune}} &0.7725&0.8327& 0.8014& 0.6312& 0.7936&0.7031\cr  
			&OmniAnomaly \cite{su2019robust} &0.7416&\textbf{0.9776}&0.8434& 0.8867& 0.9117&0.8989\cr  
			&GAN-Li \cite{li2018anomaly} &0.6710&0.8706&0.7579&0.7102& 0.8706& 0.7823\cr  
			&MAD-GAN \cite{li2019mad} &0.8049& 0.8214&0.8131& 0.8517& 0.8991& 0.8747\cr  
			&LSTM-VAE \cite{park2018multimodal} &0.8551& 0.6366&0.7298&0.5257&0.9546& 0.6780\cr \cline{1-8}
			\multirow{5}{*}{P-model}&LSTM-NDT \cite{hundman2018detecting} &0.8965& 0.8846& 0.8905&0.5934&0.5374& 0.5640\cr  
			&DAGMM \cite{zong2018deep} &0.5845& 0.9058& 0.7105&0.5412&\textbf{0.9934} & 0.7007\cr
			&MTAD-GAT \cite{zhao2020multivariate}&0.8906& 0.9123& 0.9013&0.8754& 0.9440& 0.9084\cr
			&GTA \cite{chen2021learning}& 0.8911 & 0.9176 & 0.9041 & 0.9104 &0.9117 &0.9111\cr
			&ATCN &{ \textbf{0.9539}}&{ 0.9019}&{ \textbf{0.9272}}&{ \textbf{0.9419}}&{ 0.9815}&{ \textbf{0.9613}}\cr 
			\bottomrule  
		\end{tabular}  
	\end{threeparttable}  
\end{table*}  

\subsection{Evaluation Metrics}
We evaluated the performance of various methods by the most frequently used metrics in the anomaly detection community, including Precision, Recall, and F1 score over the testing set: 
\begin{equation}
	\begin{aligned}
		{\rm Precision}&=\rm{\frac{TP}{TP+FP}}\\ {\rm Recall}&=\rm{\frac{TP}{TP+FN}}\\
		{\rm F1}&=\rm{\frac{2 \times Precision \times Recall}{Precision + Recall}}
	\end{aligned}
\end{equation}
where $\rm{TP, FP, FN}$ are the numbers of true positives, false positives, and false negatives. 
Following the evaluation strategy in \cite{su2019robust}, we use point-wise scores. In practice, anomalies at one time point in time series usually occur to form contiguous abnormal segments. Anomaly alerts can be triggered in any subset of the actual anomaly window. Therefore, the entire anomalous segment is correctly detected if any one time point is detected as anomaly by the model. The implementation of ATCN is publicly available\footnote{ The code is available at \url{ https://github.com/Lliang97/Spacecraft-Anonamly-Detection}}.
%

\subsection{Experimental Setup}
We set the historical sliding window size to 100 and predict the value at the next timestamp. Moreover, the dropout strategy was applied to prevent overfitting and dropout rate is fixed to 0.1. The models are trained using the Adam optimizer with a learning rate initialized as $1e^{-3}$ and batch size as 256 for 100 epochs. Table \ref{configuration} summarises the network configuration details. For anomaly detection on each test dataset, we apply a grid search on all possible anomaly thresholds and choose the best threshold to report F1 score. 


\subsection{Result and Analysis}
The experimental results of various anomaly detection approaches on two datasets are reported in Table \ref{tab:performance_comparison}. It can be seen that our method shows superior performance and achieves the best F1 score 0.9272 for SMAP and 0.9613 for MSL. Specifically, we can draw the following conclusions. 
\begin{itemize}
	\item Compared to the newest method GTA published in 2021, our model achieves 2.31\% and 5.02\% F1 improvement on two datasets respectively. Similarly, the precision is also boosted by 6.28\% and 3.15\% on SMAP and MSL, respectively. MTAD-GAT has comparable performance as GAT on SMAP.
	
	\item Our method surpasses OmniAnomaly and DAGMM in most cases. Though OmniAnomaly captures temporal dependencies by stochastic RNNs, it ignores the correlations between variables, which is vital for multivariate times-series anomaly detection. The performance of DAGMM is not satisfactory since it does not consider historical temporal information. Therefore, it is important to dig the correlated information in terms of variable and temporal features.
	\item LSTM-NDT creates a model for each telemetry channel, which leads to high expensive for modeling. It produces poor performance on MSL.
	\item GAN-based anomaly detection methods GAN-Li and MAD-GAN give inferior performance because they fail to fully consider the correlations between
	variables. 
	\item Most methods use RNNs to capture temporal dependency, which is limited by the need to retain historical information in memory gates, restricting the ability to model long-term sequences and being subject to long computing times. TCNs with multi-stack dilated convolution and residual connection can capture long sequence that  provides more pattern information and be calculated in parallel. Thus, our proposed model has clear advantages compared with existing models, which makes it attractive for spacecraft telemetry anomaly detection.
\end{itemize}

\begin{figure}[htp]
	\centering
	\includegraphics[width=0.45\textwidth]{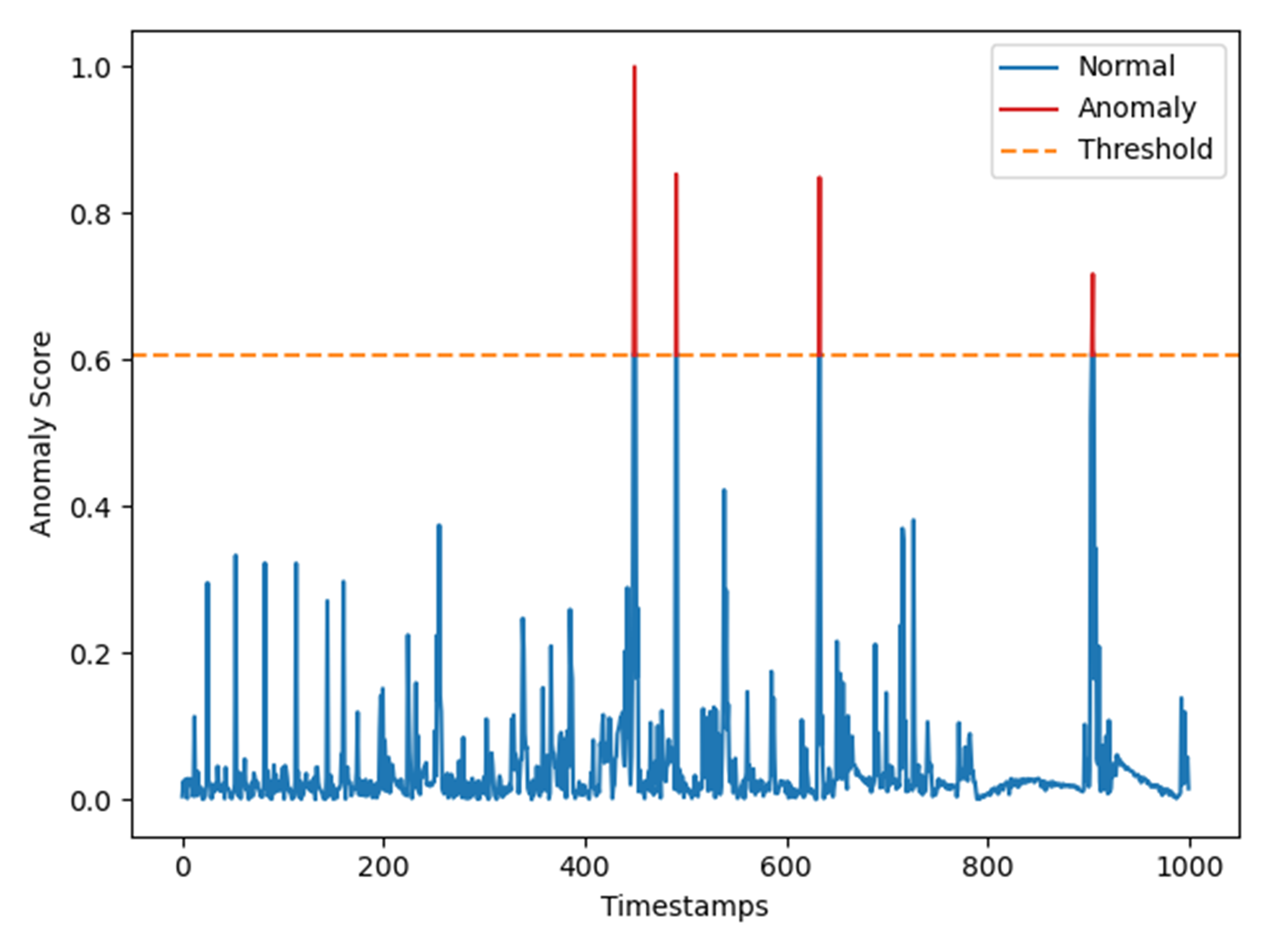}
	\caption{Examples of anomaly detection results on MSL.}
	\label{example}
\end{figure}

We give an example of anomaly detection results on MSL in Fig. \ref{example}. When spacecraft telemetry data arrive, the residuals of predicted data and actual data are calculated to obtain anomaly scores,  based on which threshold is calculated to determine whether it is an anomaly on each timestep. The blue line represents the curve of the anomaly score, the orange line denotes the calculated threshold, and the red line can be recognized as the anomalous segment. It can be seen that the detected outliers are mostly consistent with the true outliers, indicating the accuracy of our proposed anomaly detection algorithm.

\begin{figure}[htp]
	\centering
	\includegraphics[width=0.45\textwidth]{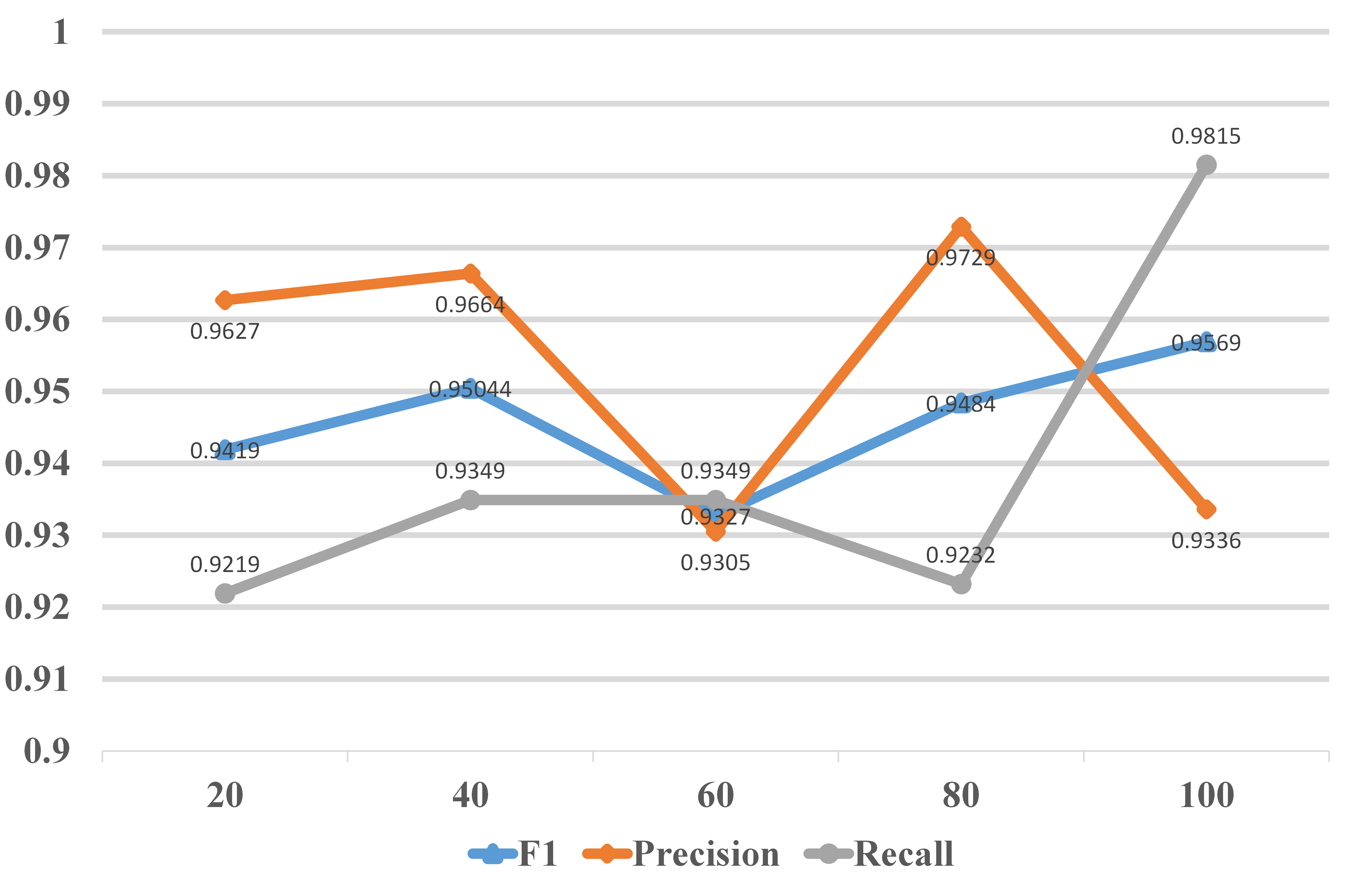}
	\caption{Influence of window size on MSL.}
	\label{Influence:window}
\end{figure}

In order to observe the influence of long-term time series on the model, we set different window sizes, i.e., $w=[20, 40, 60, 80, 100]$, to predict the value at the next timestamp. Then, the F1, precision, and recall are shown in Fig. \ref{Influence:window}.  We can observe that F1 reaches the highest value at $w=100$, so our model can capture the relationships in long time series. 

\subsection{Ablation Study}
\begin{figure}[!htbp] \footnotesize
	\begin{center}
		\begin{tabular}{cc}
			\includegraphics[width=0.98\linewidth]{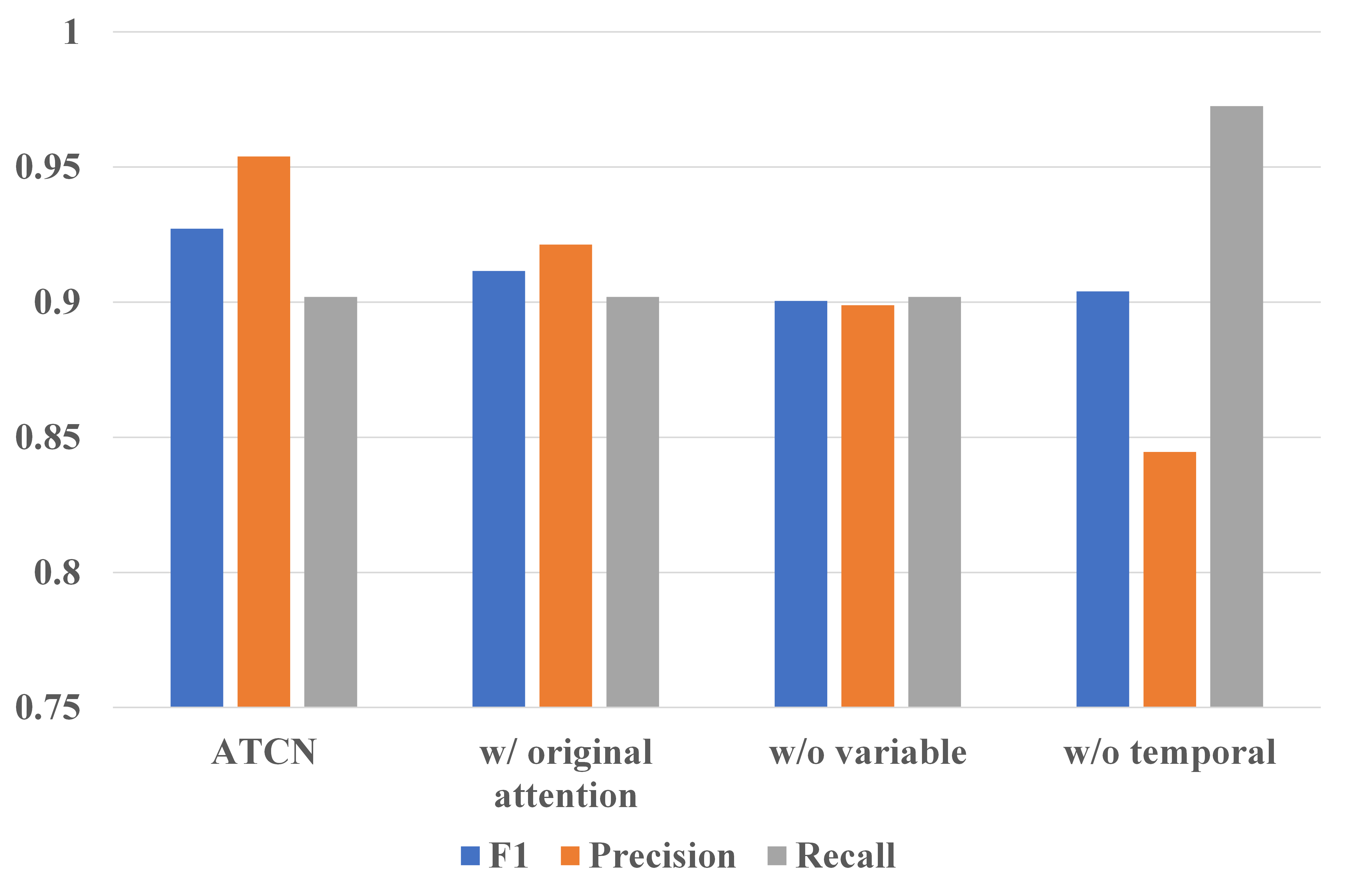} \\
			\includegraphics[width=0.98\linewidth]{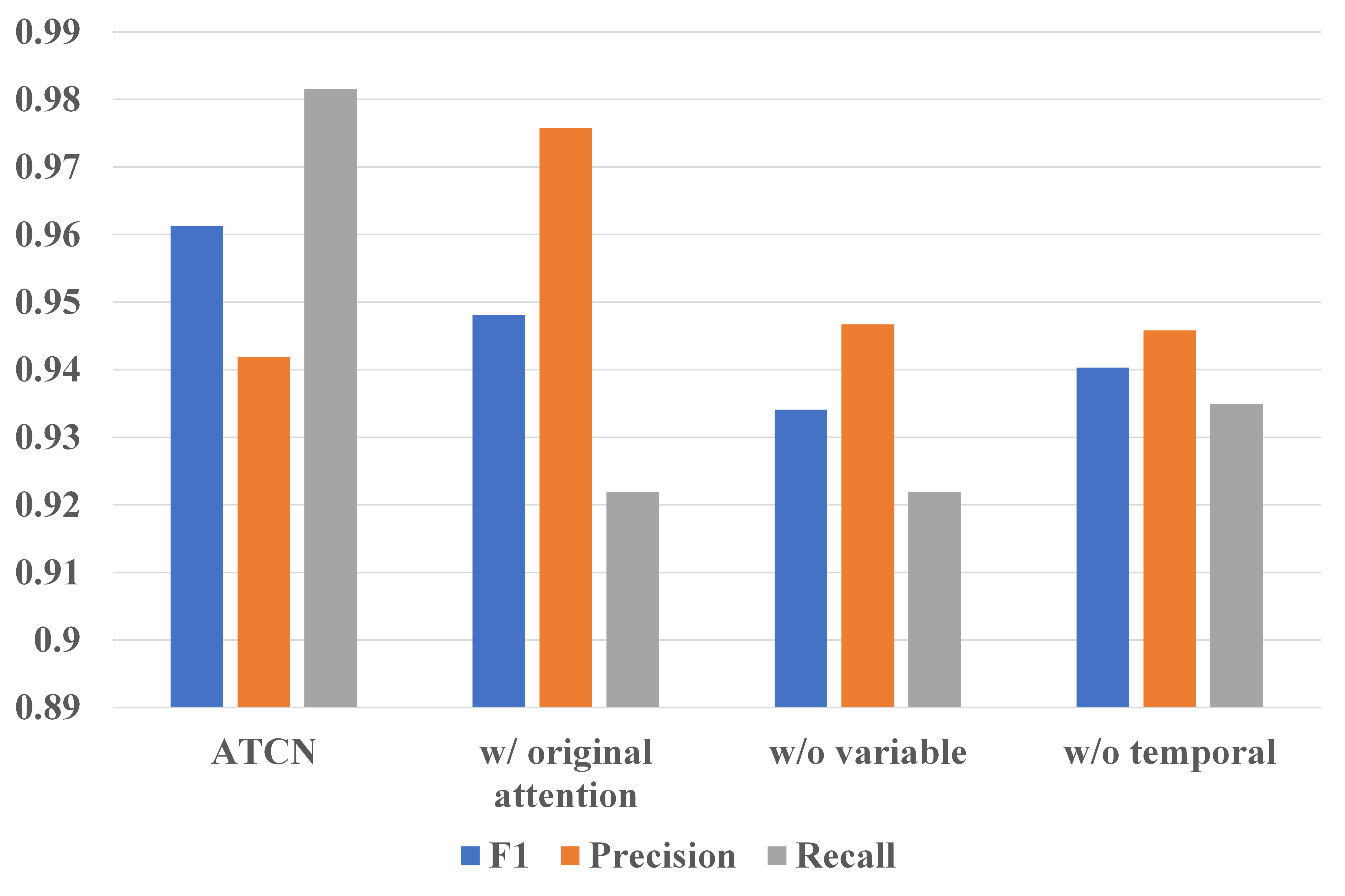} \\
		\end{tabular}
	\end{center}
	\caption{The ablation study of SMAP and MSL.}
	\label{ablation: w/o}
\end{figure}

To verify the necessity of each component in our method, we exclude variable attention, temporal attention, and the dynamic attention was replaced with original graph attention respectively to see how the model performs after these operations. 
As shown in Fig. \ref{ablation: w/o}, our original model always achieves the best F1, which verifies that the dynamic graph attention capturing correlations benefits model performance. On the other hand, the performance will degrade if the variable or temporary relationship information are removed, which validates the importance of relationship modeling in dealing with multivariate time series anomaly detection. 

Since using different thresholds will reach different results, the anomaly detection model was tested with two other statistical threshold selection methods to verify its robustness. With multivariate time series of $N$ observations, we compute an anomaly score as $S=\{{S_1, S_2,..., S_N}\}$ for every observation.
Specifically, the epsilon method \cite{hundman2018detecting} and Peak Over Threshold (POT) \cite{siffer2017anomaly} method were adopted to select threshold in offline train period. Epsilon method calculates the threshold $\epsilon$ based on error sequence $S$  according to 
\begin{equation}
\bm{\epsilon} = \mu(S)+z\sigma(S),
\end{equation}
where $\epsilon$ is determined by:
\begin{equation}
	\epsilon=argmax(\bm{\epsilon})=\frac{\Delta \mu(S) / \mu(S)
		+ \Delta \sigma(S) / \sigma(S)
	}{|S_a|+|P_{seq}|^2}
\end{equation}
where $\mu$ is expectation and $\sigma$ is standard deviation.
\begin{align*}
	\Delta \mu(S) &= \mu(S) - \mu(\{S_i \in S | S_i < \epsilon \}) \\
	\Delta \sigma(S) &= \sigma(S) - \sigma(\{S_i \in S | S_i < \epsilon \}) \\
	S_a &= \{S_i \in S | S_i > \epsilon\} \quad  P_{seq} = sequence\ of\ S_a
\end{align*}

POT is the second theorem in Extreme Value Theory (EVT) and is a statistical theory to find the law
of extreme values in a sequence. The advantage of EVT is that it does not need to assume the data distribution and learns the threshold by a generalized Pareto distribution (GPD)
with parameters. The GPD function is as follows:
\begin{equation}
	\overline{F}(s) = P(th - S  > s | S < th ) \sim \left(1 + \frac{\gamma s} {\beta}\right) ^{ - \frac{1}{\gamma}}
\end{equation}
where $th$ is the initial threshold, $\gamma$ and $\beta$ are
shape and scale parameters of GPD, and $S$ is any value in $\{{S_1, S_2,...,S_N}\}$. The portion below the threshold $th$ is denoted as $th - S$ and it is empirically set to a low quantile. Then, the values of parameters $\hat{\gamma}$ and $\hat{\beta}$ are obtained by Maximum Likelihood Estimation. The final threshold can be calculated by:
\begin{equation}
	th_{F} \simeq th - \frac{\hat{\beta}}{\hat{\gamma}}\left(\left(\frac{qN}{N_{th}}\right)^{-\hat{\gamma}}-1 \right)
\end{equation}
where $q$ is the proportion of $S < th$, $N_{th}$ denotes the number of $S < th$. 

From Table \ref{tab:different-methods}, it can be concluded that the proposed ATCN model has competitive performance with respect to above two threshold selection approaches. We observe that epsilon method has achieved very close results benefiting from appropriate modeling of outliers. The POT method does not work well in terms of F1 because it has two parameters that need to be empirically fine-tuned. 

\begin{table}[htbp]
	\centering
	\setlength{\abovecaptionskip}{0pt}%
	\setlength{\belowcaptionskip}{10pt}
	\caption{The different threshold selection methods} 
	\label{tab:different-methods}
	\renewcommand\arraystretch{1.2}
	\begin{tabular}{c|c|c|c|c}
		\hline
		\multicolumn{2}{c|}{\textbf{Datasets}} &
		Precision &
		Recall & 
		F1
		
		\\ \hline
		\multirow{3}{*}{SMAP} &ATCN & 0.9539 & 0.9019 & 0.9272 \\ \cline{2-5}
		& epsilon method & 0.9463 & 0.9019 & 0.9235 
		\\ \cline{2-5}
		&POT method& 0.9870 & 0.6285 & 0.7680  \\ \hline
		\multirow{3}{*}{MSL} &ATCN & 0.9419 & 0.9815 & 0.9613 \\ \cline{2-5}
		&epsilon method& 0.9194 & 0.9815 & 0.9494 \\ \cline{2-5}
		&POT method& 0.9844 & 0.8156 & 0.8921 
		\\ \hline
	\end{tabular}
\end{table}

\section{Conclusions}
This paper proposes a model for spacecraft anomaly detection task. A temporal convolution networks-based architecture to process multivariate time-series was designed. In particular, the dynamic graph attention module was utilized to learn the correlation between variables and times and perform weighted embedding. After that, a temporal convolution networks module was designed to learn the patterns of the spacecraft data and detect anomalies. Experiments on two real-world NASA spacecraft datasets show that our method outperforms baselines in terms of F1, Precision, and Recall. Thus, it is helpful for operator to localize and understand anomalies. In the future, the researchers plan to investigate online training model toward the employment of proposed method in future spacecraft.

%



\bibliographystyle{spmpsci}      
\bibliography{template}   

\begin{thebibliography}{10}
\providecommand{\url}[1]{{#1}}
\providecommand{\urlprefix}{URL }
\expandafter\ifx\csname urlstyle\endcsname\relax
  \providecommand{\doi}[1]{DOI~\discretionary{}{}{}#1}\else
  \providecommand{\doi}{DOI~\discretionary{}{}{}\begingroup
  \urlstyle{rm}\Url}\fi

\bibitem{zhang2019contribution}
Zhang, R., Tu, R., Fan, L., Zhang, P., Liu, J., Han, J., Lu, X.: Contribution
  analysis of inter-satellite ranging observation to bds-2 satellite orbit
  determination based on regional tracking stations.
\newblock Acta Astronautica \textbf{164}, 297--310 (2019)

\bibitem{chen2021imbalanced}
Chen, J., Pi, D., Wu, Z., Zhao, X., Pan, Y., Zhang, Q.: Imbalanced satellite
  telemetry data anomaly detection model based on bayesian lstm.
\newblock Acta Astronautica \textbf{180}, 232--242 (2021)

\bibitem{jiang2022anomaly}
Jiang, L., Xu, H., Liu, J., Shen, X., Lu, S., Shi, Z.: Anomaly detection of
  industrial multi-sensor signals based on enhanced spatiotemporal features.
\newblock Neural Computing and Applications pp. 1--13 (2022)

\bibitem{wang2021anomaly}
Wang, Y., Wu, Y., Yang, Q., Zhang, J.: Anomaly detection of spacecraft
  telemetry data using temporal convolution network.
\newblock In: 2021 IEEE International Instrumentation and Measurement
  Technology Conference (I2MTC), pp. 1--5. IEEE (2021)

\bibitem{chalapathy2019deep}
Chalapathy, R., Chawla, S.: Deep learning for anomaly detection: A survey.
\newblock arXiv preprint arXiv:1901.03407  (2019)

\bibitem{hundman2018detecting}
Hundman, K., Constantinou, V., Laporte, C., Colwell, I., Soderstrom, T.:
  Detecting spacecraft anomalies using lstms and nonparametric dynamic
  thresholding.
\newblock In: Proceedings of the 24th ACM SIGKDD international conference on
  knowledge discovery \& data mining, pp. 387--395 (2018)

\bibitem{chang1992satellite}
Chang, C.: Satellite diagnostic system: An expert system for intelsat satellite
  operations.
\newblock In: Proc. IVth European Aerospace Conference (EAC 91) (1992)

\bibitem{wang2019multivariate}
Wang, B., Liu, D., Peng, Y., Peng, X.: Multivariate regression-based fault
  detection and recovery of uav flight data.
\newblock IEEE Transactions on Instrumentation and Measurement \textbf{69}(6),
  3527--3537 (2019)

\bibitem{galeano2006outlier}
Galeano, P., Pe{\~n}a, D., Tsay, R.S.: Outlier detection in multivariate time
  series by projection pursuit.
\newblock Journal of the American Statistical Association \textbf{101}(474),
  654--669 (2006)

\bibitem{li2016anomaly}
Li, L., Hansman, R.J., Palacios, R., Welsch, R.: Anomaly detection via a
  gaussian mixture model for flight operation and safety monitoring.
\newblock Transportation Research Part C: Emerging Technologies \textbf{64},
  45--57 (2016)

\bibitem{zhang2005network}
Zhang, Y., Ge, Z., Greenberg, A., Roughan, M.: Network anomography.
\newblock In: Proceedings of the 5th ACM SIGCOMM conference on Internet
  Measurement, pp. 30--30 (2005)

\bibitem{choi2021deep}
Choi, K., Yi, J., Park, C., Yoon, S.: Deep learning for anomaly detection in
  time-series data: Review, analysis, and guidelines.
\newblock IEEE Access  (2021)

\bibitem{mathonsi2022multivariate}
Mathonsi, T., Zyl, T.L.v.: Multivariate anomaly detection based on prediction
  intervals constructed using deep learning.
\newblock Neural Computing and Applications pp. 1--15 (2022)

\bibitem{shi2022unsupervised}
Shi, Y., Shen, H.: Unsupervised anomaly detection for network traffic using
  artificial immune network.
\newblock Neural Computing and Applications pp. 1--21 (2022)

\bibitem{ding2019real}
Ding, N., Ma, H., Gao, H., Ma, Y., Tan, G.: Real-time anomaly detection based
  on long short-term memory and gaussian mixture model.
\newblock Computers \& Electrical Engineering \textbf{79}, 106458 (2019)

\bibitem{zhang2019deep}
Zhang, C., Song, D., Chen, Y., Feng, X., Lumezanu, C., Cheng, W., Ni, J., Zong,
  B., Chen, H., Chawla, N.V.: A deep neural network for unsupervised anomaly
  detection and diagnosis in multivariate time series data.
\newblock In: Proceedings of the AAAI conference on artificial intelligence,
  vol.~33, pp. 1409--1416 (2019)

\bibitem{kang2019robust}
Kang, Z., Pan, H., Hoi, S.C., Xu, Z.: Robust graph learning from noisy data.
\newblock IEEE transactions on cybernetics \textbf{50}(5), 1833--1843 (2020)

\bibitem{hsieh2019unsupervised}
Hsieh, R.J., Chou, J., Ho, C.H.: Unsupervised online anomaly detection on
  multivariate sensing time series data for smart manufacturing.
\newblock In: 2019 IEEE 12th Conference on Service-Oriented Computing and
  Applications (SOCA), pp. 90--97. IEEE (2019)

\bibitem{park2018multimodal}
Park, D., Hoshi, Y., Kemp, C.C.: A multimodal anomaly detector for
  robot-assisted feeding using an lstm-based variational autoencoder.
\newblock IEEE Robotics and Automation Letters \textbf{3}(3), 1544--1551 (2018)

\bibitem{audibert2020usad}
Audibert, J., Michiardi, P., Guyard, F., Marti, S., Zuluaga, M.A.: Usad:
  unsupervised anomaly detection on multivariate time series.
\newblock In: Proceedings of the 26th ACM SIGKDD International Conference on
  Knowledge Discovery \& Data Mining, pp. 3395--3404 (2020)

\bibitem{wen2019time}
Wen, T., Keyes, R.: Time series anomaly detection using convolutional neural
  networks and transfer learning.
\newblock In: AI for Internet of Things Workshop (2018)

\bibitem{su2019robust}
Su, Y., Zhao, Y., Niu, C., Liu, R., Sun, W., Pei, D.: Robust anomaly detection
  for multivariate time series through stochastic recurrent neural network.
\newblock In: Proceedings of the 25th ACM SIGKDD international conference on
  knowledge discovery \& data mining, pp. 2828--2837 (2019)

\bibitem{li2019mad}
Li, D., Chen, D., Jin, B., Shi, L., Goh, J., Ng, S.K.: Mad-gan: Multivariate
  anomaly detection for time series data with generative adversarial networks.
\newblock In: International Conference on Artificial Neural Networks, pp.
  703--716. Springer (2019)

\bibitem{zhou2019beatgan}
Zhou, B., Liu, S., Hooi, B., Cheng, X., Ye, J.: Beatgan: Anomalous rhythm
  detection using adversarially generated time series.
\newblock In: IJCAI, pp. 4433--4439 (2019)

\bibitem{choi2020gan}
Choi, Y., Lim, H., Choi, H., Kim, I.J.: Gan-based anomaly detection and
  localization of multivariate time series data for power plant.
\newblock In: 2020 IEEE international conference on big data and smart
  computing (BigComp), pp. 71--74. IEEE (2020)

\bibitem{chen2021learning}
Chen, Z., Chen, D., Zhang, X., Yuan, Z., Cheng, X.: Learning graph structures
  with transformer for multivariate time series anomaly detection in iot.
\newblock IEEE Internet of Things Journal \textbf{9}(12), 9179--9189 (2022)

\bibitem{meng2019spacecraft}
Meng, H., Zhang, Y., Li, Y., Zhao, H.: Spacecraft anomaly detection via
  transformer reconstruction error.
\newblock In: International Conference on Aerospace System Science and
  Engineering, pp. 351--362. Springer (2019)

\bibitem{iverson2012general}
Iverson, D.L., Martin, R., Schwabacher, M., Spirkovska, L., Taylor, W., Mackey,
  R., Castle, J.P., Baskaran, V.: General purpose data-driven monitoring for
  space operations.
\newblock Journal of Aerospace Computing, Information, and Communication
  \textbf{9}(2), 26--44 (2012)

\bibitem{zhao2020multivariate}
Zhao, H., Wang, Y., Duan, J., Huang, C., Cao, D., Tong, Y., Xu, B., Bai, J.,
  Tong, J., Zhang, Q.: Multivariate time-series anomaly detection via graph
  attention network.
\newblock In: 2020 IEEE International Conference on Data Mining (ICDM), pp.
  841--850. IEEE (2020)

\bibitem{li2018anomaly}
Li, D., Chen, D., Goh, J., Ng, S.k.: Anomaly detection with generative
  adversarial networks for multivariate time series.
\newblock In: 7th International Workshop on Big Data, Streams and Heterogeneous
  Source Mining: Algorithms, Systems, Programming Models and Applications
  (2018)

\bibitem{zong2018deep}
Zong, B., Song, Q., Min, M.R., Cheng, W., Lumezanu, C., Cho, D., Chen, H.: Deep
  autoencoding gaussian mixture model for unsupervised anomaly detection.
\newblock In: International conference on learning representations (2018)

\bibitem{bai2018empirical}
Bai, S., Kolter, J.Z., Koltun, V.: An empirical evaluation of generic
  convolutional and recurrent networks for sequence modeling.
\newblock arXiv preprint arXiv:1803.01271  (2018)

\bibitem{he2019temporal}
He, Y., Zhao, J.: Temporal convolutional networks for anomaly detection in time
  series.
\newblock In: Journal of Physics: Conference Series, vol. 1213, p. 042050. IOP
  Publishing (2019)

\bibitem{fang2022structure}
Fang, R., Wen, L., Kang, Z., Liu, J.: Structure-preserving graph representation
  learning.
\newblock In: ICDM (2022)

\bibitem{velivckovic2017graph}
Velickovic, P., Cucurull, G., Casanova, A., Romero, A., Lio, P., Bengio, Y.:
  Graph attention networks.
\newblock In: International Conference on Learning Representations (2018)

\bibitem{brody2021attentive}
Brody, S., Alon, U., Yahav, E.: How attentive are graph attention networks?
\newblock In: International Conference on Learning Representations (2022)

\bibitem{mirsky2018kitsune}
Mirsky, Y., Doitshman, T., Elovici, Y., Shabtai, A.: Kitsune: An ensemble of
  autoencoders for online network intrusion detection.
\newblock machine learning \textbf{5}, 2 (2018)

\bibitem{siffer2017anomaly}
Siffer, A., Fouque, P.A., Termier, A., Largouet, C.: Anomaly detection in
  streams with extreme value theory.
\newblock In: Proceedings of the 23rd ACM SIGKDD International Conference on
  Knowledge Discovery and Data Mining, pp. 1067--1075 (2017)

\end{thebibliography}

%
%

\end{document}